\documentclass[letterpaper, 10 pt, conference]{ieeeconf}  

\IEEEoverridecommandlockouts                              

\usepackage{times}

\usepackage{multicol}
\usepackage[bookmarks=true]{hyperref}
\usepackage{amsmath,amssymb,amsfonts}
\usepackage{booktabs}
\usepackage{caption}
\usepackage{graphicx}
\usepackage{xcolor}
\usepackage{bbm}
\usepackage{nicefrac}
\usepackage{comment}
\usepackage{algorithm}
\usepackage{algpseudocode}

\newcommand\argmin[1]{\underset{#1}{\text{argmin }}}

\newcommand{\sref}[1]{Sec. \ref{#1}}
\newcommand{\figref}[1]{Fig. \ref{#1}}
\newcommand{\tabref}[1]{Table \ref{#1}}
\newcommand{\RR}{\mathbb{R}}

\title{\LARGE \bf
Multimodal Safe Control for Human-Robot Interaction
}

\author{Ravi Pandya, Tianhao Wei, Changliu Liu
\thanks{Authors are with Carnegie Mellon University, Pittsburgh, PA, USA \tt\small \{rapandya, twei2, cliu6\}@andrew.cmu.edu}
}

\begin{document}

\maketitle
\thispagestyle{plain}
\pagestyle{plain}

\begin{abstract}
Generating safe behaviors for autonomous systems is important as they continue to be deployed in the real world, especially around people. In this work, we focus on developing a novel safe controller for systems where there are multiple sources of uncertainty. We formulate a novel multimodal safe control method, called the Multimodal Safe Set Algorithm (MMSSA) for the case where the agent has uncertainty over which discrete mode the system is in, and each mode itself contains additional uncertainty. To our knowledge, this is the first energy-function-based safe control method applied to systems with multimodal uncertainty. We apply our controller to a simulated human-robot interaction where the robot is uncertain of the human's true intention and each potential intention has its own additional uncertainty associated with it, since the human is not a perfectly rational actor. We compare our proposed safe controller to existing safe control methods and find that it does not impede the system performance (i.e. efficiency) while also improving the safety of the system.
\end{abstract}

\IEEEpeerreviewmaketitle

\section{Introduction}
Guaranteeing that autonomous systems can stay safe is becoming increasingly important as robots are being deployed in the real world. Such autonomous agents need to consider how to keep themselves safe (e.g. avoid collisions with objects) but also how to keep other agents safe. In particular, robots need to ensure that people working with and around them are also kept safe from harm. This means we need to design autonomous robots in a way that we know they will not put people in unsafe situations.

One well-known approach to this problem is to design a safe control law that guarantees that the the safe region of the state space is \textit{forward invariant}, i.e. the system will not leave the safe region once it enters it. Additionally, many safe controllers are designed to guarantee \textit{finite-time convergence} to the safe region if the system is in an unsafe state. Many safe control laws are designed with a scalar \textit{energy function}, designed such that if the energy function is negative, the system is in a safe state. A safe control law can then be designed such that if the system is in an unsafe state, the energy function will decrease (i.e. the time derivative of the energy function is negative). Some examples of energy-function-based safe control are control barrier functions (CBFs) \cite{ames2019control}, sliding mode control \cite{gomez2022safe} and the safe set algorithm (SSA) \cite{liu2014control}. 

\begin{figure}
    \centering
    \includegraphics[width=0.85\columnwidth]{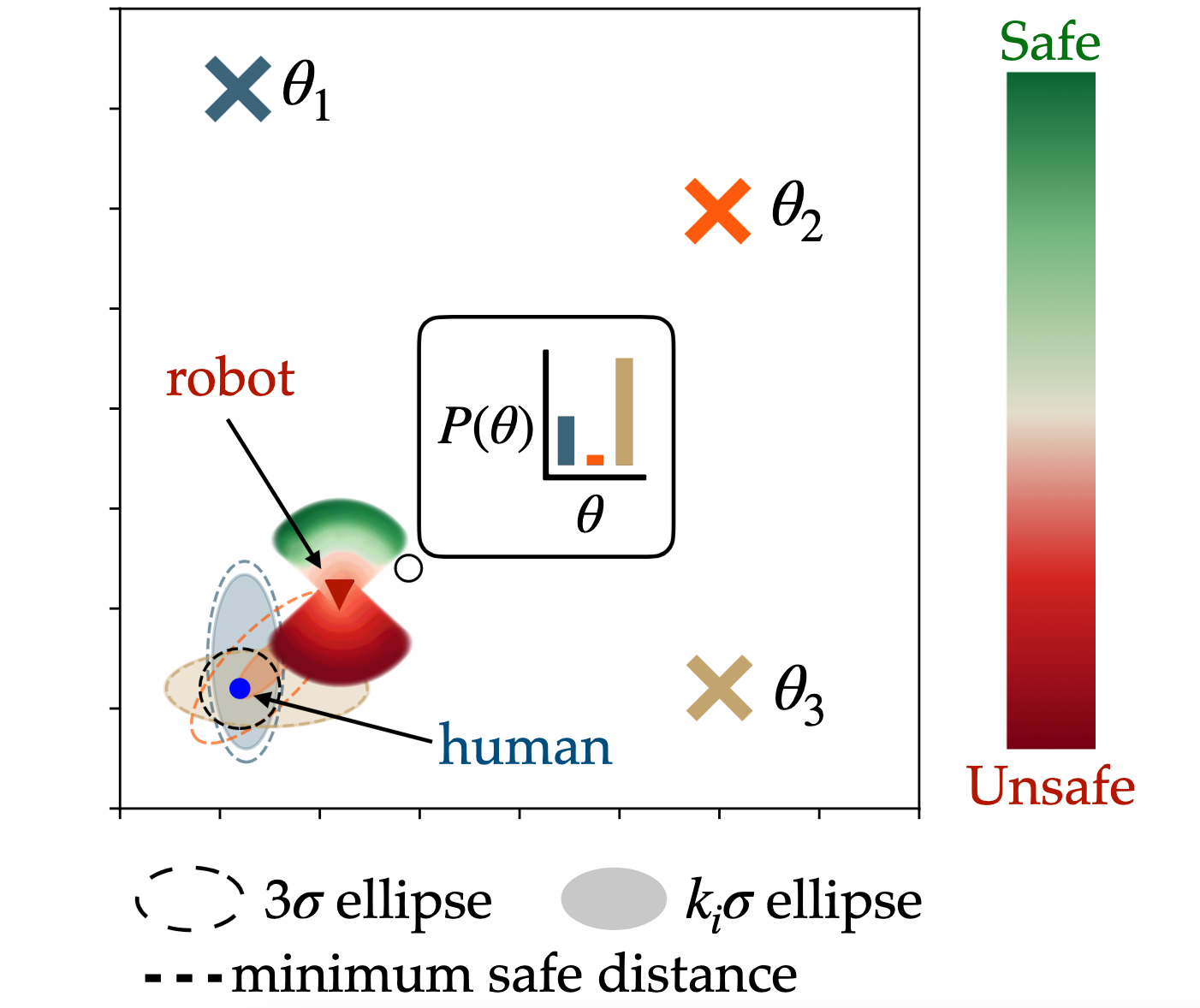}
    \caption{Depiction of a multimodal human-robot system where the robot needs to stay safe with all potential goals $\theta_i$ the human might choose. The shaded region shows how safe potential future positions are, colored by the distance to the safe set of the corresponding control action.}
    \label{fig:front_figure}
    \vspace{-0.2in}
\end{figure}

We focus on the application of safe control methods to human-robot interaction (HRI), so we need to additionally handle safe control under uncertainty since we cannot perfectly predict how humans act. It has long been known in fields like cognitive science and behavioral economics that humans are not perfectly rational decision makers \cite{simon1997models, conlisk1996bounded}, so we should model human behavior with probabilities. Much recent work in HRI has focused on inferring a human's intention probabilistically \cite{sadigh2016planning, bestick2018learning}, and recently researchers have trained models for probabilistic trajectory forecasting of humans \cite{salzmann2020trajectron++, tolstaya2021identifying}. These models all have multimodal outputs, recognizing that from the same state, different humans may have different intentions. These methods are generally focused on prediction accuracy and do not deal explicitly with safety, let alone safety with respect to the full multimodal distribution over human actions. 

In our work, we introduce a new safe control formulation to deal with the \textit{multimodality} of human actions where the robot explicitly accounts for the human's intention while computing its safe control.
There is existing work on energy-function-based safe control under uncertainty \cite{liu2015safe, wei2022persistently}, though they only consider unimodal uncertainty. 

Our contributions in this work are the following:
\begin{enumerate}
    \item A novel formulation of multimodal safe control for a robot interacting with people;
    \item A derivation of a least-conservative safe control law for a dynamics model with multimodal uncertainty;
    \item A comparison with existing safe control.
\end{enumerate}

\section{Problem Formulation}
\subsection{Energy-Function-Based Safe Control}
\label{sec:efb_safety}
As previously mentioned, a popular approach for designing a safe control law is to construct a scalar energy function (or \textit{safety index}) that indicates how safe the system state currently is. Concretely, the safety index $\phi: \mathcal{X}\rightarrow \RR$ maps the state space $\mathcal{X}$ to the reals. The function should be designed such that the system is safe if $\phi(x) \leq 0$. We want to design a safe control law that decreases $\phi(x)$ when the system is unsafe, and even in safe situations, $\phi(x)$ should not increase too quickly. This means our safety constraint becomes an inequality constraint on $\dot\phi(x)$:
\begin{align}
    \dot\phi(x) = \nabla_x\phi^\top(x)\dot x \leq \eta(\phi)\label{eqn:phi_dot_const}
\end{align}
where $\eta(\phi)$ is our margin for safety, or the maximum rate of increase of $\phi(x)$. For ease of notation going forward, we will write $\nabla_x\phi^\top(x)$ simply as $\nabla_x\phi^\top$.

The general problem of designing $\phi(x)$ for an arbitrary system is challenging, but is not the focus of this work. There is a breadth of existing literature on writing or synthesizing safety indices \cite{liu2014control, clark2021verification, zhao2023safety}. Thus we assume that we are given a safety index $\phi$ for our system that we must design a safe control law around. In particular, we build on the Safe Set Algorithms, a CBF-style approach for computing safe controls for a nonlinear control-affine system.

\subsection{Multimodal Dynamical System}
We start by defining a nonlinear system in control-affine form\footnote{An arbitrary control system can be transformed to the control-affine form through dynamic extension by introducing a new virtual control input as the derivative of the existing control input.}:
\begin{equation}
    \dot x = m(x,\theta) + g(x) u,
    \label{eqn:mean_dynamics}
\end{equation}
where the state is $x\in\RR^{n}$, the control is $u\in\RR^{m}$, $m(x,\theta): \RR^{n}\rightarrow\RR^{n}$, $g(x): \RR^{m}\rightarrow\RR^{n}$ and $\theta$ denotes the discrete system mode. We call this dynamical system the \textit{mean} or \textit{noiseless} dynamics. We are given a safety index $\phi$, so---following from \eqref{eqn:phi_dot_const}---our safety constraint for the noiseless system is:
\begin{align}
    \nabla_x\phi^\top(m(x,\theta) + g(x) u) \leq \eta(\phi)&\\
    \nabla_x\phi^\top m(x,\theta) + \nabla\phi^\top(x)g(x)u \leq \eta(\phi)&\\
    \underbrace{\nabla_x\phi^\top g(x)}_{L(x)}u \leq \underbrace{\eta(\phi) - \nabla_x\phi^\top m(x,\theta)}_{S(x,\theta)}& \label{eqn:mm_const_mean}.
\end{align}

Ultimately, this gives us a half-space constraint on the set of control inputs, so our set of safe controls is defined as $\mathcal{U}^S(x)=\{u \mid L(x)u \leq S(x,\theta)\}$.

We now introduce the dynamical system that includes multimodal uncertainty. We ultimately wish to do safe control with respect to this system, which previous methods have not addressed. This \textit{uncertain} system with multimodal uncertainty has dynamics:
\begin{equation}
    \dot x = f(x,\theta) + g(x) u, 
    \label{eqn:dynamics}
\end{equation}
where $\theta$ follows a discrete probability distribution $\theta\sim\{\theta_1,\theta_2\ldots\theta_N\}$ and $f(x,\theta)\sim \mathcal{N}(m(x,\theta), \Sigma(x,\theta))$ is normally distributed centered at the mean dynamics \eqref{eqn:mean_dynamics}. We call this uncertainty in $f$ \textit{additive} uncertainty. This is contrasted with uncertainty in $g$, which we call \textit{multiplicative} uncertainty, because of the way the uncertainty enters into the dynamics relative to the control input $u$. In this work, we do not consider any uncertainty in $g$. 
In \sref{sec:hri_dynamics} we will introduce the human-robot instantiation of this dynamical system, where it is natural to represent uncertainty in the human's dynamics in $f$ because the robot cannot exert direct control over the human.

Following the same procedure as with the noiseless dynamics, we can write down the safety constraint with the uncertain system \eqref{eqn:dynamics}:
\begin{equation}
    \underbrace{\nabla_x\phi^\top g(x)}_{L(x)}u \leq \underbrace{\eta(\phi) - \nabla_x\phi^\top f(x,\theta)}_{\bar S(x,\theta)}
    \label{eqn:mm_const}
\end{equation}

Note that $L(x)$ is the same as the original constraint. Since $f(x,\theta_i)$ is drawn from an unbounded probability distribution (Gaussian) as previously noted, this constraint is not possible to satisfy, because $\bar S(x)$ cannot be computed. Instead, we introduce a chance constraint because want to satisfy \eqref{eqn:mm_const} constraint with high probability:
\begin{equation}
    P\left[\dot\phi \leq \eta(\phi)\right] \geq 1 - \epsilon,
    \label{eqn:safety_level}
\end{equation}
where $1-\epsilon$ is our desired probability of safety. 

In the following section, we will describe our novel procedure for deriving a safe controller that can satisfy this multimodal chance constraint.

\section{Multimodal Safe Set Algorithm}
\label{sec:mmssa}
\begin{figure*}
    \centering
    \includegraphics[width=0.9\textwidth]{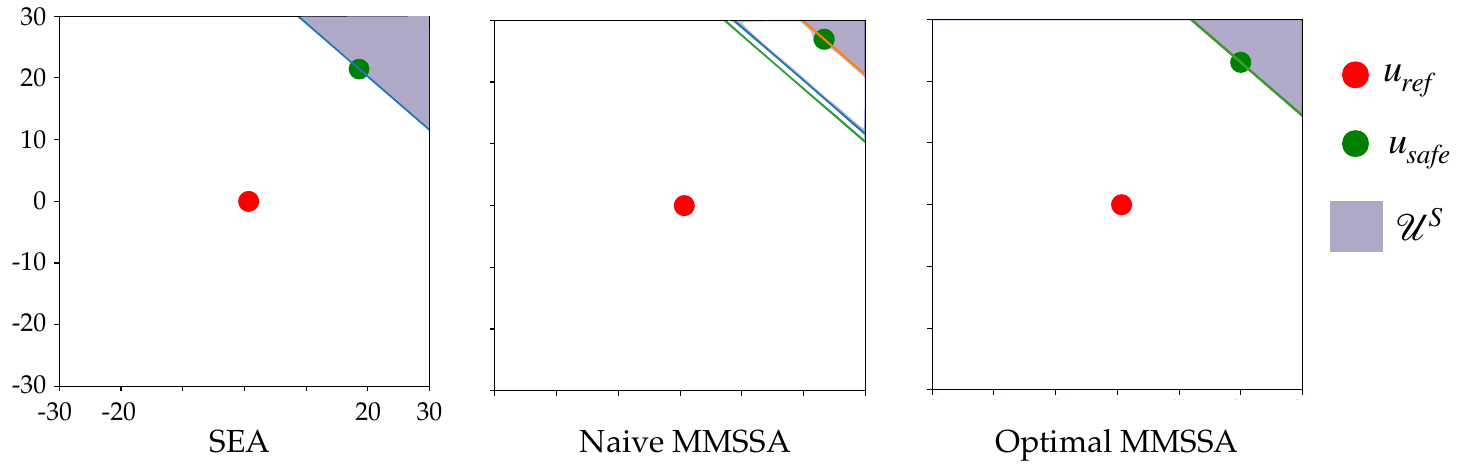}
    \caption{The control constraints from each safe control method at the same system state. }
    \label{fig:control_constraints}
    \vspace{-0.2in}
\end{figure*}

\subsection{Naive Multimodal Safe Control (N-MMSSA)}
\label{sec:n-mmssa}
Each mode $\theta_i$ has some probability of being the \textit{true} mode that the system is in at each timestep, denoted by $P(\theta_i)$. Since we are working with normally distributed $f(x,\theta_i)$, the original probabilistic constraint will be satisfied if 
\begin{equation}
    \sum_i P(\theta_i) \chi_n^2(k_i^2) \geq 1-\epsilon
    \label{eqn:gauss_safety}
\end{equation}
where $\chi_n^2$ is the CDF of the Chi-squared probability distribution with $n$ degrees of freedom and $k_i$ is the Mahalanobis distance (generalized standard deviation) away from the center of the Gaussian we are trying to bound for mode $i$.
For example, in the 1D case, if the system is unimodal and  $\epsilon=0.003$, then we can choose $k=3$ to satisfy the probabilistic constraint because $\chi^2_1(9)=0.997$, meaning 99.7\% of the probability mass is within a Mahalanobis distance of 3 (i.e. the standard 3$\sigma$ bound).

Let us consider the difference $\rho(x,\theta)$ between the RHS of the noiseless constraint $S(x,\theta)$ and the RHS of the uncertain constraint $\bar S(x,\theta)$:
\begin{align}
    &\rho(x,\theta) = S(x,\theta) - \bar S(x,\theta)\\
    &= \left[\eta(\phi) - \nabla_x\phi^\top m(x,\theta)\right] - \left[\eta(\phi) - \nabla_x\phi^\top f(x,\theta)\right]\\
    &= -\nabla_x\phi^\top\left[m(x,\theta) + f(x,\theta)\right]\\
    &= \nabla_x\phi^\top\left[f(x,\theta) - m(x,\theta)\right].
\end{align}

The key idea in our derivation is that \textbf{if we can bound the difference $\rho(x,\theta) = S(x,\theta)-\bar S(x,\theta)$, we can compute a safety constraint for the multimodal uncertain system}. 
Finding the maximum value of $\rho(x,\theta)$ will allow us find a \textit{computable} safety constraint $\bar S(x,\theta)=S(x,\theta)-\rho(x,\theta)$ by subtracting the maximum value of $\rho(x,\theta)$ from the original noiseless $S(x,\theta)$. For ease of notation, let $\delta(x,\theta)=f(x,\theta)-m(x,\theta)$.
The maximum value of $\rho(x,\theta)$, which we call $\gamma(x,\theta)$ can be written as:
\begin{equation}
    \gamma(x,\theta) = \max_{\delta^\top (x,\theta)\Sigma^{-1}(x,\theta)\delta(x,\theta)\leq 1} \nabla_x\phi^\top \delta(x,\theta).
\end{equation}
Concretely, $\gamma(x,\theta)$ represents the maximum value of $\rho(x,\theta)$ if we assume that $f(x,\theta)$ lies within the ellipsoid of Mahalanobis distance of $1$ from the mean $m(x,\theta)$. We call this case the $1\sigma$ worst case value of $\rho(x,\theta)$. If we assume $f(x,\theta)$ is within the ellipsoid of Mahalanobis distance of $k$ from $m(x,\theta)$, then the maximum value of $\rho(x,\theta)$ is $k\gamma(x,\theta)$. We call this the $k\sigma$ worst case value of $\rho(x,\theta)$.


We can then compute our robust control constraint that satisfies our chance constraint \eqref{eqn:safety_level} by subtracting the $k\sigma$ worst case value of $\rho(x,\theta)$ from $S(x,\theta)$.
We then get one constraint per mode that we can directly compute:
\begin{equation}
\begin{split}
    \nabla_x\phi^\top g(x) u \leq \eta(\phi) - \nabla_x\phi^\top m(x,\theta_i) -k_i\gamma(x,\theta_i)&,\\ 
    \forall i=\{1,\ldots,N\}&
    \label{eqn:safety_constraint}
\end{split}
\end{equation}
where $k_i$ represents the desired safety level for mode $i$. To satisfy the probabilistic constraint \eqref{eqn:gauss_safety}, we need a way to pick $k_i$ bound for each mode. One easy way to guarantee this constraint is satisfied is to pick a fixed value $k$ for all $k_i$ such that $\chi_n^2(k^2)\geq1-\epsilon$. For example, our experiments use $\epsilon=0.003$, so for a 1D system, any $k\geq3$ will satisfy the constraint. We refer to this method as the Naive Multimodal Safe Set Algorithm (\textbf{N-MMSSA}). The final constraints for the naive version are then: 

\begin{equation}
\begin{split}
    \underbrace{\nabla_x\phi^\top g(x)}_{L(x)} u \leq \underbrace{\eta(\phi) - \nabla_x\phi^\top m(x,\theta_i) - k\gamma(x,\theta_i)}_{S_i(x,\theta_i)}&,\\ 
    \forall i=\{1,\ldots,N\}&.
\end{split}
\end{equation}
where $k$ is chosen such that $\chi_n^2(k^2)\geq1-\epsilon$. These constraints define the set of safe controls $\mathcal{U}^S(x)=\{u \mid L(x) u \leq S_i(x,\theta_i), \forall i\}$. Since all the constraints are parallel in the control space (since $L(x)$ does not depend on $\theta$), we can take the most conservative constraint to define the safe set of controls:
\begin{equation}
\begin{split}
    \mathcal{U}^S = \{u \mid L(x) u\leq S^*(x)\}&\\
    S^*(x) = \min_{i} S_i(x,\theta_i)&
\end{split}
\label{eqn:qp_constraint}
\end{equation}

The safe control input will be determined by projecting a reference control $u_{ref}$ to the set of safe controls $\mathcal{U}^S$. Since $L(x)u\leq S^*(x)$ is linear in $u$, this can be formulated as a quadratic program:
\begin{equation}
    u_{safe} = \argmin{u\in \mathcal{U}^S} \parallel u - u_{ref}\parallel_2^2.
    \label{eqn:safe_qp}
\end{equation}
Here, the constraint $u\in\mathcal{U}^S$ is linear based on $\mathcal{U}^S$ defined in \eqref{eqn:qp_constraint}. Following the original Safe Set Algorithm \cite{liu2014control}, the safety controller should only be active when necessary:
\begin{equation}
    u=
    \begin{cases}
      u_{safe}, & \text{if}\ \phi\geq 0 \\
      u_{ref}, & \text{otherwise}
    \end{cases}
    \label{eqn:final_controller}
\end{equation}

Though this method will satisfy the probabilistic safety constraint, it may be overly conservative, since it stays safe with respect to a fixed bound per mode $\theta_i$, even if the probability $P(\theta_i)$ that the system is in that mode is arbitrarily low. Notice that in \figref{fig:control_constraints}, N-MMSSA picks the most conservative control constraint, so we look for a way to more intelligently choose $k$ for each mode.

\subsection{Optimal Multimodal Safe Control (O-MMSSA)}
\label{sec:o-mmssa}
Instead of naively choosing a fixed $k\sigma$ bound for all modes, the least-conservative safe control will try to distribute the constraints equally among all modes. We will define $\mu(x,\theta_i,k_i)=\nabla_x\phi^\top m(x,\theta_i)+k_i\gamma(x,\theta_i)$, so we can choose the desired $\mathbf{k}=[k_1,\ldots,k_N]^\top$ by solving the following constrained optimization:
\begin{equation}
    \begin{split}
        &\min_{\mathbf{k}} \left\{\max_{i,j} \{\mu(x,\theta_i,k_i) - \mu(x,\theta_j,k_j)\} + ||\mathbf{k}||_2\right\}\\
        &\text{s.t. } \sum_i P(\theta_i)\chi_n^2(k_i^2) \geq 1-\epsilon.
    \end{split}
\end{equation}


This optimization is trying to find the smallest $k\sigma$ bound for each mode that results in the smallest possible difference between all pairs of control constraints while satisfying the probabilistic constraint \eqref{eqn:gauss_safety}. After optimizing for all $k_i$'s, we can then directly compute the safety constraints shown in \eqref{eqn:safety_constraint}. We can see in \figref{fig:control_constraints} that all three resulting control constraints are essentially overlapping. In practice, the inner optimization can be computed exactly since it is over discrete variables, and the outer optimization is solved with Sequential Least Squares Programming (SLSQP) \cite{kraft1988software}. This controller resulting from this method of optimally selecting the $k\sigma$ bound per mode is called the Optimal Multimodal Safe Set Algorithm (\textbf{O-MMSSA}). The final control input can be computed using the same method as in N-MMSSA, shown in \eqref{eqn:final_controller}.


\subsection{Comparison with Unimodal Safe Control (SEA)}
\label{sec:sea}
As a baseline comparison, we use a controller from prior work that does safe control under Gaussian uncertainty \cite{liu2015safe}, called SEA (Safe Exploration Algorithm). In the original paper, the robot considers only the most-likely mode $\theta'=\text{argmax}_{\theta}P(\theta)$ of the human. Just as in the original paper, the robot considers staying safe with respect to a fixed Mahalanobis distance to satisfy the desired level of safety for $\epsilon=0.003$ \eqref{eqn:safety_level}. 

We note that this approach is not guaranteed to be safe with respect to the multimodal system, since this \textit{unimodal} approach essentially takes $k_i=3$ for one mode and $k_i=0$ for all other modes. In a simple 1D bimodal system where $P(\theta_1)=P(\theta_2)=0.5$, it's easy to see that simply setting $k_1=3,k_2=0$ will not satisfy \eqref{eqn:gauss_safety}. Intuitively, we can see in \figref{fig:control_constraints} that SEA picks the least conservative constraint in this case, which may not be safe if the true system is in a different mode.

We additionally note that the original paper that proposes SEA modifies the safety index $\phi$ in order to stay safe with respect to a $3\sigma$ bound, but we instead include it in the control constraint for a more fair comparison between the methods discussed here (so all methods use the same safety index). This means the robot's safe control imposes the following linear constraint on the control space:
\begin{equation}
    \nabla_x\phi^\top g(x) u \leq \eta(\phi) - \nabla_x\phi^\top m(x,\theta')-k'\gamma(x,\theta')
\end{equation}
where $\chi_n^2(k'^2)\geq1-\epsilon$. Once again, the final control input can by computed by solving the same QP as in N-MMSSA and O-MMSSA \eqref{eqn:final_controller}.

\section{Application to Human-Robot System}
\subsection{Dynamics}
\label{sec:hri_dynamics}
Though the formulation itself is applicable to general multimodal control-affine systems, we focus on the case of human-robot interaction, since this is a realistic real-world instantiation of the multimodal safe control problem. Specifically, we consider the joint state $x=[x_R^\top, x_H^\top]^\top$ and $\dot x$ to be the joint human-robot system dynamics:
\begin{align}
    \dot x &= f(x,\theta) &&+ g(x)u \\
    \begin{bmatrix}\dot x_R \\ \dot x_H\end{bmatrix} &= \begin{bmatrix}f_R(x_R) \\ h(x_H,x_R,\theta)\end{bmatrix} &&+ \begin{bmatrix}g_R(x_R) \\ 0\end{bmatrix} u_R.
\end{align}
where $x_H\in\RR^{n_H}$ and $x_R\in\RR^{n_R}$ are the states of the human and robot respectively, $u_R\in\RR^{m_R}$ is the robot's control input, $f_R(x_R) : \RR^{n_R}\rightarrow\RR^{n_R}$ and $g_R(x_R): \RR^{n_R} \rightarrow \RR^{n_R\times m_R}$ define the robot's dynamics and $h(x_H,x_R,\theta) : \RR^{n_H}\times\RR^{n_R} \rightarrow \RR^{n_H}$ defines the human's dynamics. Since the robot's dynamics are deterministic, we have 
\begin{align}
    m(x,\theta) &= \begin{bmatrix}f_R(x_R) \\ m_H(x_H,x_R,\theta)\end{bmatrix}\\
    \Sigma(x,\theta) &= \begin{bmatrix}0 & 0\\ 0 & \Sigma_H(x,\theta)\end{bmatrix}
\end{align}
where $\Sigma_H(x,\theta)$ defines the covariance matrix of the human's dynamics $h(x_H,x_R,\theta)\sim\mathcal{N}(m_H(x_H,x_R,\theta), \Sigma_H(x,\theta))$.  Note that the robot has no direct control over the human. In the human-robot system, each $\theta_i$ corresponds to a potential intention of the human (for example, different goal locations to reach). The human's true intention $\theta^*$ is unknown to the robot, so the robot needs to estimate it online. 

\subsection{Human Intent Inference}
Our multimodal safe control method uses the assumption that each mode $\theta_i$ has some probability of being the true mode of the system (\sref{sec:n-mmssa}). As a result, it is natural to instantiate this in a human-robot system as the human's intent (unknown to the robot). The robot keeps a belief $b_R^t(\theta) = [P(\theta_1),\ldots,P(\theta_N)]$, which represents the probability that $\theta_i=\theta^*$. Our safe control method does not assume any particular structure for either the semantic meaning of the human's intent $\theta$ nor for how the belief is computed. It only requires that at each timestep, the robot has access to a discrete probability distribution $b_R^t(\theta)$.

Much work has been done on human intent inference, often framing the problem as inverse reinforcement learning \cite{ng2000algorithms, inga2017individual, sadigh2016planning, fisac2018probabilistically} where the intention $\theta$ represents the reward function that the human is assumed to be optimizing. Recently, many have focused on human trajectory forecasting  \cite{salzmann2020trajectron++, shi2022motion, tolstaya2021identifying} where researchers use large datasets to learn predictors that output probabilities over discrete modes. These modes could be treated as the possible intentions $\theta$ of the human. 

For simplicity and a proof of concept for our proposed multimodal safe controller, we choose a relatively simple intent estimator (still grounded in prior work) over a fixed set of goal states, as described later in \sref{sec:goal_inference}. We do not claim to introduce any novel intent estimation method.

\section{Simulation Setup}
\subsection{Simulated Dynamics Models}
The robot has deterministic unicycle dynamics $\dot x_R = f_R(x_R) + g_R(x_R)u_R$:
\begin{equation}
    \dot x_R = \begin{bmatrix}\dot r_x \\ \dot r_y \\ \dot v_R \\ \dot \psi_R\end{bmatrix} = \begin{bmatrix}v_R \cos\psi_R \\ v_R\sin\psi_R\\ 0 \\ 0\end{bmatrix} + \begin{bmatrix}0 & 0\\0 & 0\\ 1 & 0\\0 & 1\end{bmatrix}u_R
\end{equation}
where the robot's control input is the linear acceleration and steering angle velocity $u_R=[\dot v_R, \dot\psi_R]^\top$.

The human has noisy double-integrator LTI dynamics $\dot x_H = Ax_H + Bu_H + w(t)$:
\begin{equation}
    \dot x_H = \begin{bmatrix}\dot h_x \\ \dot v_x \\ \dot h_y \\ \dot v_y\end{bmatrix} = \begin{bmatrix}0 & 1 & 0 & 0\\0 & 0 & 0 & 0\\0 & 0 & 0 & 1\\0 & 0 & 0 & 0\\\end{bmatrix}x_H + \begin{bmatrix}0 & 0\\1 & 0\\ 0 & 0\\0 & 1\end{bmatrix}u_H + w(t),
\end{equation}
where $w(t)\sim\mathcal{N}(0,\Sigma)$. $m_H(x_H,x_R,\theta)$ is the dynamics without noise: $m_H(x_H,x_R,\theta) = Ax_H + Bu_H$. We assume the human's control has the following form:
\begin{equation}
    u_H = -K(x_H-\theta) + \frac{\gamma}{d^2}(C_H x_H - C_R x_R)
\end{equation}
where $C_H$ and $C_R$ are constant matrices that select the $x$ and $y$ positions of the respective agent's state, $d=|| C_R x_R - C_H x_H ||_2$ and $\gamma\in\RR$ determines how strongly the human is repelled from the robot. Specifically,
\begin{align}
    C_H &= \begin{bmatrix}
        1 & 0 & 0 & 0\\
        0 & 0 & 1 & 0
    \end{bmatrix}\\
    C_R &= \begin{bmatrix}
        1 & 0 & 0 & 0\\
        0 & 1 & 0 & 0
    \end{bmatrix}.
\end{align}

The gain $K$ is the solution to the infinite-horizon linear-quadratic regulator (LQR) problem with diagonal weight matrices $Q,R$ that minimizes
\begin{equation}
    \int_0^{\infty} \left( (x_H-\theta)^\top Q(x_H-\theta) + u_H^\top Ru_H\right) dt.
\end{equation}

\subsection{Goal Inference on Simulated Human}
\label{sec:goal_inference}
\begin{figure}
    \centering
    \includegraphics[width=0.9\columnwidth]{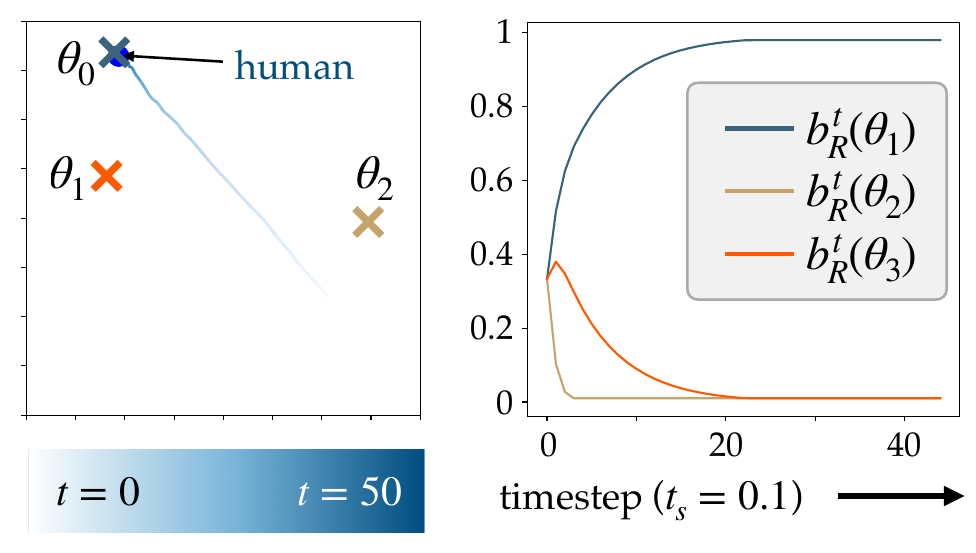}
    \caption{Bayesian inference of the simulated human's goal}
    \label{fig:bayes_inf}
    \vspace{-0.2in}
\end{figure}

In our simulations, the robot assumes a Boltzmann-rational model of the human's behavior, a model commonly used in fields related to human decision making \cite{mcfadden1973conditional, mcfadden1974measurement}, especially human-robot interaction \cite{baker2007goal, levine2012continuous}. Specifically, this means the human is exponentially more likely to choose an action if it has a higher Q-value (where the Q-value is the action-value function in the human's Markov Decision Process):
\begin{equation}
    p(u_H^t \mid x_H^t, x_R^t; \theta) = \frac{e^{\beta_R Q_H(x_H^t, x_R^t, u_H^t;\theta)}}{\int_{u_H^{'}} e^{\beta_R Q_H(x_H^t, x_R^t, u_H^{'};\theta)} },
\end{equation}
where $\beta_R$ is the rationality coefficient (sometimes called the ``inverse temperature'' parameter or the ``model confidence'').

In general, this quantity can be difficult to compute, especially computing the integral in the denominator without discretizing the action space. While this is a common approach, we instead follow prior work \cite{tian2023towards} to utilize the known form of the human's control input to compute an exact form for this probability.

The robot uses this likelihood function to update $b_R^t$ at some discrete time interval using Bayes' Rule:
\begin{equation}
\begin{split}
    &b_R^t(\theta_i) := p(\theta_i \mid x_H^{0:t}, x_R^{0:t}, u_H^{0:t}) \\
    &= \frac{p(u_H^{t} \mid x_H^{t} , x_R^{t} ; \theta_i) p(\theta_i \mid x_H^{0:t-1}, x_R^{0:t-1}, u_H^{0:t-1})}{\sum_{\theta^{'}} p(u_H^{t}  \mid x_H^{t} , x_R^{t} ; \theta^{'}) p(\theta^{'} \mid x_H^{0:t-1}, x_R^{0:t-1}, u_H^{0:t-1})}
\end{split}
\label{eqn:bayes_rule}
\end{equation}

The belief $b_R^t(\theta)$ is treated as the prior distribution for the next timestep $t+1$.

In the simulated system, each $\theta_i$ is a potential goal location that the human is moving towards. Since the human's control input is primarily governed by this goal-reaching controller computed via LQR, we use the LQR cost to define the robot's inference. In order to write down a Q-value function in closed form, the robot treats the effect of its own state on the human's control as noise in the human's decision making\footnote{This could alternatively be viewed as the robot only having knowledge of the human's goal-seeking control, so it has no choice but to include other influences on the human's control as unobserved noise.}. Drawing on prior work \cite{tian2023towards}, we consider the human's reward function to be the negative instantaneous LQR cost:
\begin{equation}
    r_H(x_H,u_H;\theta) = -(x_H-\theta)^\top Q(x_H-\theta) - u_H^\top Ru_H. 
\end{equation}

Since there is no dependence on $x_R$, we have dropped it as an argument. The Q-value (or action-value) function is only well-defined in discrete time\footnote{In continuous time, the human's action shows up as the time derivative of the optimal cost-to-go, but for simplicity, we build on existing discrete-time Bayesian inference approaches.}, so to properly do inference on the human's goal, we consider the equivalent discrete-time dynamics model of the human $(\tilde{A},\tilde{B})$. In this case, we know that the action-value function is the negative infinite-time optimal cost-to-to for LQR after the human takes $u_H$:
\begin{equation}
    Q_H(x_H, u_H; \theta) = r(x_H,u_H;\theta) - (x'-\theta)^\top P(x'-\theta)
\end{equation}
where $x'=\tilde{A}x_H+\tilde{B}u_H$ and $P$ is the solution to the discrete-time algebraic Ricatti equation (DARE): 
\begin{equation}
    P=\tilde{A}^\top P\tilde{A}-\tilde{A}^\top P\tilde{B}(R+\tilde{B}^\top P\tilde{B})^{-1}\tilde{B}^\top P\tilde{A}+Q.
\end{equation}

\begin{figure*}
    \centering
    \includegraphics[width=0.9\textwidth]{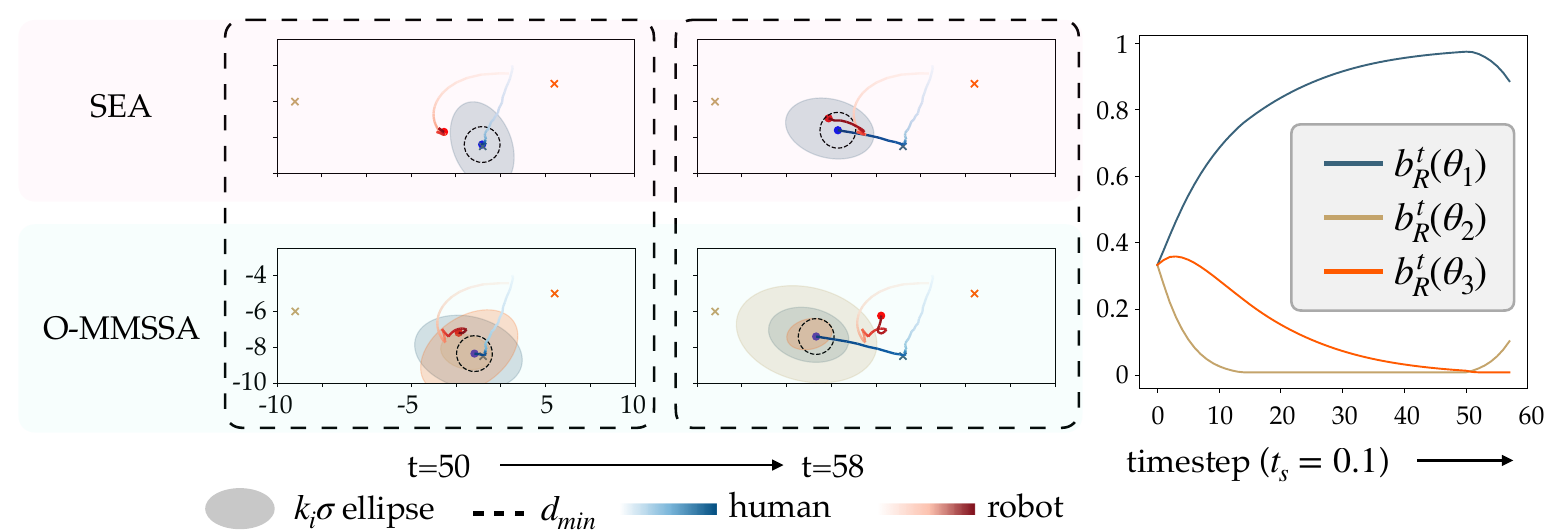}
    \caption{An example of unimodal safe control (\sref{sec:sea}) violating the minimum safe distance from the human while the proposed multimodal safe controller (\sref{sec:o-mmssa}) does not.}
    \label{fig:sea_violation}
    \vspace{-0.2in}
\end{figure*}

From the derivation in \cite{tian2023towards}, we can compute an explicit form of the denominator of $p(u_H^t \mid x_H^t; \theta)$ because it takes the form of a Guassian integral:
\begin{equation}
\begin{split}
    &\int e^{\beta_R Q_H(x_H,u_H;\theta)} \\
    &= e^{-\beta_R(x_H-\theta)^\top P(x_H-\theta)}\sqrt{\frac{(2\pi)^m}{\text{det}(2\beta_R R+2\beta_R\tilde{B}^\top P\tilde{B})}}.
\end{split}
\end{equation}

The robot then uses this computed likelihood function to update its belief using Bayes Rule \eqref{eqn:bayes_rule} given a new observation $(x_H^t, u_H^t)$.

Rolling this inference out on the simulated system, we can see in \figref{fig:bayes_inf} that the robot's belief correctly converges to the human's intended goal $\theta_0$ after about 25 observations. We can also see that the probability of $\theta_1$ initially increases, which makes sense, since the robot observes the human moving in the direction of both $\theta_0$ and $\theta_1$. Similarly, the probability of $\theta_2$ immediately decreases since the human is moving away from that goal.

\subsection{Nominal Robot Control}
\label{sec:nominal_control}
The robot's nominal (or reference) control $u_{ref}=[\dot v_R, \dot\psi_R]^\top$ drives it towards its own goal state $[g_x, g_y]$ which defines an $(x,y)$ position. Let $\delta_x=(r_x-g_x)$ and $\delta_y=(r_y-g_y)$, then:
\begin{align}
    \dot v_R = -[\delta_x\cos\psi_R + \delta_y\sin\psi_R] - k_v v_R\\
    \dot\psi_R = k_{\psi}\left[\arctan\frac{\delta_y}{\delta_x} - \psi_R\right]
\end{align}
where $k_v,k_{\psi}\in \mathbb{R}^+$ are constants. The controller is designed according to the Lyapunov function $V=\delta_x^2+\delta_y^2+v_R^2$. 

\subsection{Safe Control}
The safety index is designed as $\phi=d_{min}^2-d^2-k_{\phi}\dot d$ where $d_{min}$ is the minimum distance we want the agents to stay apart for safety, $d$ is the relative Cartesian distance between the agents and $k_{\phi}\in\mathbb{R}^+$ is a constant coefficient. In order to compute the control input for all methods, we need to compute $\nabla_{x_H} \phi$ and $\nabla_{x_R} \phi$. The derivations are left out for space, but we have:
\begin{align}
&\begin{split}
    &\nabla_{x_H} \phi = \\
    &2d_p^\top C_H + k_{\phi}\left[\frac{1}{d}d_v^\top C_H +\frac{1}{d}d_p^\top B^\top - \frac{1}{d^3}d_p^\top d_v d_p^\top C_H\right]
\end{split}\\
&\begin{split}
    &\nabla_{x_R} \phi = \\
    &-2d_p^\top C_R - k_{\phi}\left[\frac{1}{d}d_v^\top C_R + \frac{1}{d} d_p^\top \mathcal{V} -\frac{1}{d^3}d_p^\top d_v d_p^\top C_R\right]
\end{split}
\end{align}
where $d_p=C_R x_R - C_H x_H$ so that $d=||d_p||$, $d_v = \dot{d_p}$ and 
\begin{equation}
    \mathcal{V} := \begin{bmatrix}0 & 0 & \cos\theta_R & -v_R\sin\theta_R\\ 0 & 0 & \sin\theta_R & v_R\cos\theta_R\end{bmatrix}.
\end{equation}


\begin{table*}
    \centering
    \begin{tabular}[width=\columnwidth]{@{}lllll@{}}
    \toprule
    & Reference Control $u_{ref}$ & SEA  & N-MMSSA  & O-MMSSA  \\ 
    \midrule 
    safety rate (\%) & $95.0\%$ & $99.7\%$ & $\mathbf{99.8\%}$ & $99.7\%$ \\ \midrule
    \# goals reached per rollout & $12.0\pm 0.5$ & $\mathbf{11.3\pm 0.8}$  & $9.85\pm 1.1$ & $10.7\pm 1.0$ \\ \midrule
    control set size ($rad\cdot m / s^2$) & $3600.0\pm 0.0$ & $\mathbf{1491.2\pm 309.6}$  & $766.8\pm 148.8$ & $1244.4\pm 220.0$ \\ \bottomrule
    \end{tabular}
    \caption{\label{tab:sim_metrics} \textbf{[Goal-Seeking Robot]} Average safety and efficiency metrics.}
\end{table*}

\section{Simulation Results}

\begin{table*}
    \centering
    \begin{tabular}[width=\columnwidth]{@{}lllll@{}}
    \toprule
    & Reference Control $u_{ref}$ & SEA  & N-MMSSA  & O-MMSSA  \\ 
    \midrule 
    safety rate (\%) & $21.8\%$ & $90.9\%$ & $\mathbf{98.1\%}$ & $97.3\%$ \\ \midrule
    \# goals reached per rollout & 0 & $0$  & $0$ & $0$ \\ \midrule
    control set size ($rad\cdot m / s^2$) & $3600.0\pm 0.0$ & $\mathbf{1288.8\pm 178.0}$  & $652.4\pm 126.8$ & $822.8\pm 144.0$ \\ \bottomrule
    \end{tabular}
    \caption{\label{tab:sim_metrics_follow} \textbf{[Follower Robot]} Average safety and efficiency metrics.}
\end{table*}

\subsection{Goa-Seeking Robot Controller}
\label{sec:goal_seeking}
We compare the performance of our proposed controller O-MMSSA (\sref{sec:o-mmssa}) against the two baselines: N-MMSSA (\sref{sec:n-mmssa}) and SEA (\sref{sec:sea}). We are interested in two metrics: the \textit{safety} and \textit{efficiency} of each controller. In our evaluations, we set $\epsilon=0.003$ and $d_{min}=1m$. The agents' dynamics are implemented in continuous-time and integrated forward with a Runge-Kutta method (RK4) with a sampling time $t_s=0.1$ seconds. The time horizon for each trajectory is $25s$. Each simulation consists of a randomly sampled initial condition for the human and robot's positions as well as the set of goals $\{\theta_1,\ldots,\theta_3\}$.

We measure safety by looking at the percent of timesteps that the robot violated the minimum distance constraint. We measure efficiency by looking at two things: 1) the number of goals reached by the robot in the time horizon and 2) the volume of the safe set of controls (assuming $||u_R||_{\infty}\leq 30$). 

We run 100 initial conditions per controller and average the results across all (all controllers use the same set of initial conditions). The results are shown in \tabref{tab:sim_metrics} \textbf{[Goal-Seeking Robot]}. We can see that the O-MMSSA and SEA controllers reach the most goals on average, while N-MMSSA lets the robot reach fewer. In terms of control set size, SEA keeps the safe set of controls the largest, followed by O-MMSSA and finally N-MMSSA. This is a good sign, since it means our proposed controller allows the robot to be competitive with the baseline unimodal controller on task performance. We also see that N-MMSSA has the highest safety rate, as we expected (though all controllers have a safety rate above our desired 99.7\% threshold, which is a function of how often the two agents need to stay in close proximity during the task). 
While SEA may be efficient, using a unimodal safe control approach on a multimodal system may be unsafe.

However, our quantitative results still show little absolute difference between the different safe control methods on the metrics of both safety and efficiency. This is because safety-critical scenarios in this simulation are a long-tail event, as we can see by looking at the ``Reference Control'' column in \tabref{tab:sim_metrics}---the goal-seeking reference controller already keeps the system safe $95\%$ of the time. We thus note that safety is highly context-dependent so we measure the performance of our safe controller using a different nominal control where safety-critical situations occur more often.

\subsection{Human-Following Robot Controller}
In this version, the robot uses the same setup as in \sref{sec:goal_seeking}. It also uses the same nominal controller as in \sref{sec:nominal_control}, but the goal state $[g_x,g_y]$ is set to be the human's current $xy-$position $[h_x,h_y]$. In these simulations, we set $\epsilon=0.03$. The results are shown in \tabref{tab:sim_metrics_follow} \textbf{[Follower Robot]}. 

Now, we can look at the performance on the same metrics on safety and efficiency for these controllers. In terms of efficiency, the robot is not trying to reach a particular goal state, so we can only compare efficiency in terms of the size of the control set; the pattern we saw in the previous case still holds true. SEA is the least conservative, followed by O-MMSSA and finally N-MMSSA.

In terms of safety, we can see that the reference controller only keeps the system safe $21.8\%$ of the time, which means having a good safe controller is critical in this scenario. We now see that the unimodal controller (SEA) does make significant improvements over the reference control, but its 90\% safety is much less than the desired level of $97\%$. The naive multimodal method (N-MMSSA) keeps the system the most safe, while O-MMSSA keeps the system above the desired safety threshold. We can thus see empirically that our proposed safe controller (O-MMSSA) is a least-conservative safe controller, since it keeps the system above the desired safety threshold while not being overly restrictive in the feasible safe control set size.

To get a sense for why SEA may lead the system to unsafe states, we can look at \figref{fig:sea_violation}. At $t=58$, the most likely goal of the human is $\theta_1$, even though the human started moving to $\theta_2$, since the robot's intent inference takes multiple observations to change drastically\footnote{This is desired behavior in real HRI systems. Since humans are noisy actors, we don't want to infer the human's intent from just one action.}. As a result, SEA does not prepare for the possibility that the human's true goal may be different, resulting in the robot being in a state where a safety violation is inevitable, and happens at $t=58$. Meanwhile, O-MMSSA accounts for this possibility at $t=50$ and is already out of the way of the human by $t=58$. 

In both \tabref{tab:sim_metrics} and \tabref{tab:sim_metrics_follow} we see that SEA keeps the safe set of controls the largest, followed by O-MMSSA then N-MMSSA. However, since SEA results in the most safety violations, this tells us that the safety constraints are sometimes inaccurate. Meanwhile, O-MMSSA keeps the safe set of controls larger than N-MMSSA does, which means it is less conservative while still staying safe.

\section{Conclusion and Future Work}
We presented a novel formulation of safe control for systems with multimodal uncertainty where the system may be in one of multiple discrete modes and each mode has its own associated dynamics noise. We derived a least-restrictive controller (O-MMSSA) that balances the resulting control constraints from each mode to keep the safe set of controls as large as possible. We evaluated our proposed controller against a naive multimodal safe control baseline (N-MMSSA) and a unimodal safe controller (SEA) based on prior work. When evaluating on a simulated human-robot system, we found that our proposed controller keeps the human-robot system more safe that the unimodal baseline and keeps the set of safe controls larger than the naive multimodal baseline. This tells us that our proposed controller is helpful for improving both the safety and efficiency of human-robot interactions. Future work may additionally study how to deal with uncertainty in $g(x)$, as well as test how these controllers perform on physical robots interacting with real humans. 

\section*{Acknowledgments}
This material is based upon work supported by the National Science Foundation Graduate Research Fellowship under Grant No. DGE1745016 and DGE2140739 and additionally under Grant No. 2144489. Any opinions, findings, and conclusions or recommendations expressed in this material are those of the authors and do not necessarily reflect the views of the National Science Foundation. This work is additionally supported by the Manufacturing Futures Institute, Carnegie Mellon University, through a grant from the Richard King Mellon Foundation.

\bibliographystyle{IEEEtran}
\bibliography{references}

\end{document}